\documentclass[12pt,a4paper]{icicel}

\def\currentvolume{14}
\def\currentissue{7}
\def\currentyear{2020}
\def\currentmonth{July}
\def\ppages{653–659}

\usepackage[dvips]{graphicx}
\usepackage{epsfig}
\usepackage{epsf}

\usepackage{colortbl}
\usepackage{tabularx}
\usepackage{rotating}
 \usepackage{algorithm2e}
\newcolumntype{b}{>{\hsize=1.4\hsize}X}
\newcolumntype{s}{>{\hsize=.4\hsize}X}
\newcolumntype{g}{>{\hsize=.9\hsize}X}
\usepackage{float}
\usepackage{graphicx}
\usepackage{amssymb}
\usepackage{url}
\usepackage{url}
\usepackage{subcaption}

\usepackage{amsmath}
\usepackage{amssymb}
\usepackage{array}
\newcolumntype{C}{>{\centering\arraybackslash}p{4.9em}}
\usepackage{amsthm}
\theoremstyle{definition}

\begin{document}


\title[ICIC Express Letters  ]
      {Predicting Seminal Quality with the Dominance-Based Rough Sets Approach}

\author[N.Dehouche]{}

\maketitle

\centerline{\scshape  Nassim Dehouche$^1$}
 \medskip
{\footnotesize
\centerline{$^1$Business Administration Division}
\centerline{Mahidol University international College} 
\centerline{SalayaL, 73170, Thailand}
\centerline{nassim.deh@mahidol.edu} }
\medskip
\medskip

\centerline{Received November 2019; accepted February 2020
}

%%\centerline{(Communicated by xxxx)}

\medskip



\begin{abstract}
The paper relies on the clinical data of a previously published study. We identify two very questionable assumptions of said work, namely confusing
evidence of absence and absence of evidence, and neglecting the ordinal nature of attributes‘ domains. We then show that using an adequate ordinal methodology such as the dominance-based rough sets approach (DRSA) can significantly improve the predictive accuracy of the expert system, resulting in almost complete accuracy for a dataset of 100 instances. Beyond the performance of DRSA in solving the diagnosis problem at hand, these results suggest the inadequacy and triviality of the underlying dataset. We provide links to open data from the UCI machine learning repository to allow for an easy verification/refutation of the claims made in this paper.\\

\end{abstract}


\section{Introduction}
Reports of a global decline in male fertility and declining sperm counts \cite{mishra} have engendered significant academic interest in investigating the causes of this public health challenge. Among numerous approaches to undertake this challenge,  artificial intelligence techniques offer promising decision support to clinicians in the detection of male fertility issues. In this regard, the classification of human sperm morphometry based on set standards has been a very successful line of research. Indeed, the visual appearance of sperm has been shown to correlate to male fertility potential \cite{shape} and automatic image processing techniques \cite{morpho} can detect abnormal sperm shapes. An exhaustive study in \cite{gold} compared four supervised learning methods (1- Nearest Neighbor, naive Bayes, decision trees and Support Vector Machine (SVM)) and three shape-based descriptors (Hu moments, Zernike moments and Fourier descriptors) for this task, finding that the best classification performance was achieved by the Fourier descriptor and SVM. More recently, deep learning techniques \cite{deep, deep2} have been successfully applied to the same problem.\\

Another fruitful line of research aims at the early detection of male fertility issues based on lifestyle factor, which can indeed increase the success rate of treatment. However, despite their promises and predictive power, the performance of this approach is highly dependent on the quality and representativeness of the collected data. Thus, the present paper intends to highlight some existing limitations in the measurement of this aspect. We rely on the clinical data of \cite{gil}, referred to as the "Assisted Reproduction" dataset,  which were made publicly available on the UCI Machine Learning repository \cite{UCI}, by the first author of that publication, and used as a reference dataset by many studies \cite{zhou, big, derp}. 

\section{Data and Previous Results}
\begin{center}
\begin{figure}[!htbp]

\includegraphics[width=\textwidth]{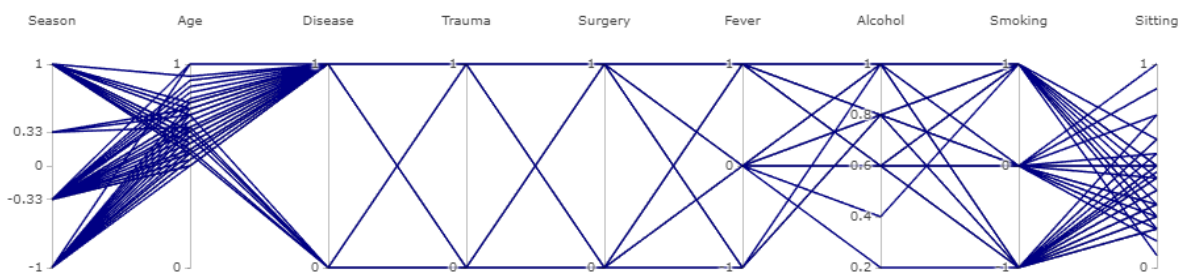}
\caption{\label{normal}Parallel coordinates of the attribute values presented by \textit{Normal} cases}
\end{figure}
\end{center}

\begin{center}
\begin{figure}[!htbp]

\includegraphics[width=\textwidth]{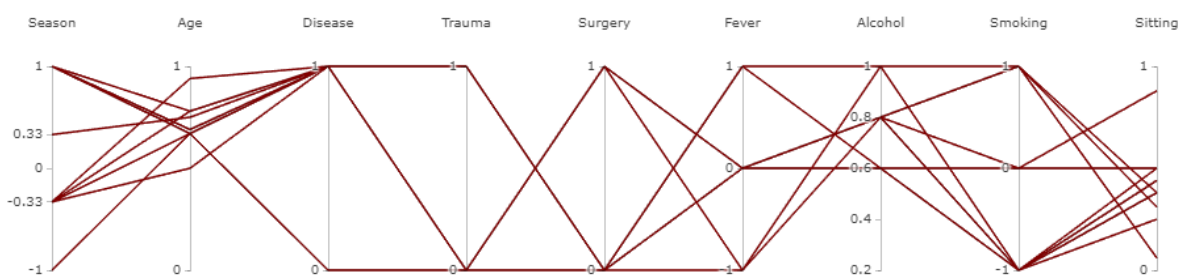}
\caption{\label{altered} Parallel coordinates of the attribute values presented by \textit{Altered} cases}
\end{figure}
\end{center}

The dataset records 9 attributes pertaining to the lifestyle habits, socio-demographic and environmental factors and health status of 100 volunteers aged 18 to 36 years, who provided a semen sample analyzed according to the WHO 2010 criteria \cite{WHO}. Based on this analysis, volunteers were classified into two classes \textit{normal} (N), or \textit{altered} (O), based on the sperm concentration they present. Table \ref{attributes} lists and describes the attributes characterizing each volunteer. Figure \ref{normal} and Figure \ref{altered} respectively present the attribute values presented by  \textit{normal} and \textit{altered} cases, in parallel coordinates form.

This binary classification problem consists in predicting the \textit{Output}, given the values of attributes \textit{Season}, \textit{Age}, \textit{Disease}, \textit{Trauma}, \textit{Surgery}, \textit{Fever}, \textit{Alcohol}, \textit{Smoking} and \textit{Sitting}. 

To address this problem, the authors compare the performance of three Artificial Intelligence methods, Decision Trees (DT), MultiLayer Perception (MLP) and Support Vector Machines (SVM), using the following classical performance indicators based on the numbers of True Positives (TP), True Negatives (TN), False Positives (FP) and False Negatives (FN) in the classification of the available 100 cases.
\begin{itemize}
\item Classification accuracy (\%)= $\frac{TP +TN}{TP+FP+FN+TN}\times  100$
\item Sensitivity (\%)= $\frac{TP}{TP+FN}\times 100$ 
\item Specificity (\%)= $\frac{TN}{FP+TN}\times 100$ 
\item Positive predictive value $(\%)=\frac{TP}{TP+FP}\times 100$
\item Negative predictive value (\%)=$\frac{TN}{FN+TN}\times 100$
\end{itemize}
\begin{center}
\begin{table}[!htbp]
\caption{\label{attributes}List of attributes with their initial domains and normalized values}
\resizebox{\columnwidth}{!}{
\begin{tabular}{|llll|} 
\cellcolor[gray]{0.8} \textbf{Attribute}& \multicolumn{1}{p{7cm}}{\cellcolor[gray]{0.8} \textbf{Description}}& \multicolumn{1}{p{4cm}}{\cellcolor[gray]{0.8} \textbf{Domain}}& \multicolumn{1}{p{4cm}|}{\cellcolor[gray]{0.8}\textbf{Normalization}}\\
\hline

Season &\multicolumn{1}{p{7cm}}{Season in which the analysis was performed}&   \multicolumn{1}{p{4cm}}{\{Winter, Spring, Summer, Fall\}}& \multicolumn{1}{p{4cm}|}{\{-1, -0.33, 0.33, 1\}}\\
\hline
Age &\multicolumn{1}{p{7cm}}{Age of the volunteer at the time of analysis}&   \multicolumn{1}{p{4cm}}{[18,36]}&\multicolumn{1}{p{4cm}|}{[0,1]}\\
\hline
Disease & \multicolumn{1}{p{7cm}}{Childish diseases (i.e., chicken pox, measles, mumps or polio)}&\multicolumn{1}{p{4cm}}{\{Yes, No\}}&\multicolumn{1}{p{4cm}|}{\{0, 1\}}\\

\hline
Trauma & \multicolumn{1}{p{7cm}}{Accidents or serious trauma}&\multicolumn{1}{p{4cm}}{\{Yes, No\}}&\multicolumn{1}{p{4cm}|}{\{0, 1\}}\\
\hline
Surgery & \multicolumn{1}{p{7cm}}{Surgical interventions}&\multicolumn{1}{p{4cm}}{\{Yes, No\}}&\multicolumn{1}{p{4cm}|}{\{0, 1\}}\\
\hline
Fever & \multicolumn{1}{p{7cm}}{High fevers in the last year}&   \multicolumn{1}{p{4cm}}{\{Less than three months ago, More than three months ago, No\}}&\multicolumn{1}{p{4cm}|}{\{-1, 0, 1\}}\\
\hline
Alcohol & \multicolumn{1}{p{7cm}}{Frequency of alcohol consumption}&   \multicolumn{1}{p{4cm}}{\{Several times a day, Every day, Several times a week, Once a week, Hardly ever or never\}} &\multicolumn{1}{p{4cm}|}{\{0.2, 0.4, 0.6, 0.8, 1\}}\\
\hline
Smoking & \multicolumn{1}{p{7cm}}{Smoking habit}&   \multicolumn{1}{p{4cm}}{\{Never, Occasionally, Daily\}}&\multicolumn{1}{p{4cm}|}{\{-1, 0, 1\}}\\
\hline
Sitting & \multicolumn{1}{p{7cm}}{Number of hours spent sitting per day}&   \multicolumn{1}{p{4cm}}{[0,16]}&\multicolumn{1}{p{4cm}|}{[0,1]}\\
\hline
Output & \multicolumn{1}{p{7cm}}{Diagnosis}&  \multicolumn{1}{p{4cm}}{\{Normal, Altered\}}&\multicolumn{1}{p{4cm}|}{\{N, O\}}\\
\hline
\end{tabular}

}

\end{table}
\end{center}


Results for the three methods, as they appear in \cite{gil} are presented in Table \ref{results}.

\section{Literature Review}
We discuss two inconsistent assumptions in \cite{gil}, namely confusing evidence of absence and absence of evidence (the former do not appear in environmental and lifestyle factors) and neglecting ordinal properties of attributes' domains. Thus the first questionable assumption of \cite{gil} lies in learning symmetrically from "Normal" and "Altered" cases. Indeed, there is a difference in nature between these two classes, in that the presence and interactions of the considered environmental factors can cause an Altered output. However, the absence of these environmental factors does not “cause”, nor explain a Normal output. Consequently, we propose that learning should focus on the the "Altered" class, the "Normal" class being considered a default class. Secondly, the environmental, lifestyle and occupational factors modeled by the classification attributes considered in \cite{gil} are known to negatively affect male fertility, which is admittedly the reason why they were considered in the survey of volunteers in that study. From a decision theoretic perspective, this means that attributes and classes are of an ordinal nature \cite{bloch} and thus classification should be monotonic \cite{mono}. For instance, all other factors being equal a patient who consumes alcohol more frequently cannot generate a "Normal" output, when a patient who consumes less alcohol generates an "Altered" output.

More formally, for data of an ordinal nature, the monotonicity requirement \cite{arie, cano} states that given two objects $a$ and $b$ to be classified, if $a$ presents values that are no worse than those presented by $b$, on each attribute, then a classification system should assign $a$ to a class than is ranked at least as high as the class $a$ is assigned. For the classification problem at hand, this property means that for the same season of analysis, the output of formally consistent classification system, for an older subject $b$ who would present more severe values for childish disease, trauma, surgery, fever, alcohol consumption, smoking and would sit longer than a subject $b$, cannot not be "Normal", when the output for $b$ is "Altered". The three classification methods considered in \cite{gil} (DT, MLP and SVM) do not ensure that this common sense property is satisfied. In addition to ensuring the formal consistency of a classification system when the input and output are of an ordinal nature, ensuring that a classification system satisfies the monotonicity requirement facilitates the detection of inconsistencies among learning examples and substantially reduces the number of classification rules \cite{arie}.

\section{Proposed Approach}

The Dominance-Based Rough Set Approach (DRSA) is an extension of the classical Rough Theory introduced by Pawlak \cite{pawlak} that explicitly gives consideration to attributes of an ordinal nature \cite{drsa}. This approach is based on the dominance binary relation and computes two sets, known as the \textit{upward} and \textit{downward} unions associated for each class of the output. \\

The \textit{upward union} associated with a class is composed of said class and all classes ranked higher, when the \textit{downward union} the considered class and all classed ranked lower. Similarly, given a learning case $a$, the \textit{dominating set} associated with it is defined as the set of all learning cases whose values on all attributes are at least as high as those of $a$. Finally, the \textit{dominating set} associated with $a$ is the set of all learning cases that do not present any value that is higher than that presented by $a$, on any attribute. \\

Logical rules induced through the DRSA aim at approximating the \textit{upward} or \textit{downward unions} of classes and have a classical "\text{If} (conditions) \text{Then} (Output)" form, in which (Conditions) is a conjunction of elementary conditions in the form of lower or upper bounds on the attributes, and (Output) is an assignment to an \textit{upward} \textit{downward unions} of classes. For the \textit{upward union} $Cl^{\geq}$ (resp. the \textit{downward union}  $Cl^{\leq}$) of an output class  $Cl^{\leq}$, the induced logical rules would suggest that an object satisfying their corresponding logical conditions should be assigned  \textit{at least} (resp.  \textit{at most}) to class $Cl_t$. \\

Finding a rule base $G$ that would exhaustively cover all learning cases, with a minimum number of rules is known to be an NP-Hard problem \cite{andersen}. The DOMLEM algorithm \cite{drsa} aims at minimizing this number heuristically. Given a set of attributes $F$, let us denote by $F^{\prime}\subseteq F$ a subset of attributes over which the elementary conditions of rules are stated, $E$ denotes a conjunction of elementary conditions $e$, that is candidate to constituting the condition part of an elementary condition, while $[E]$ is the notation for a set of cases covered by $E$. In the DOMLEM algorithm $E$ would be accepted as the condition part of a rule, if and only if $\cap_{e\in E}[e]\subseteq B$, where $B$ is an upward or downward union of classes considered as input. The choice of elementary conditions $e$ that would become part of conjunction $E$ is based on the evaluation of $E\cup\{e\}$ by a function denoted $Evaluation( )$. Several versions of this function may be used.\\

The version of the algorithm used here chooses the elementary rule providing the largest ratio  $\frac{|[E\cup\{e\}]\cap G|}{|[E]\cup\{e\}|}$, in a strategy that consists in covering the maximum number of cases with the minimum number of elementary conditions. An alternative strategy would, for instance, aim at choosing the elementary rule $e$ that minimizes the number of currently uncovered cases verifying it. To ensure minimality rules are checked iteratively and redundant elementary conditions and rules are removed from the final rule base.\\

As previously stated, for the problem at hand, we consider the "Normal" class to be a default class and focus learning on the downward union associated with cases from the "Altered" class (that is the class itself). Table \ref{rules} presents the resulting rule base of nine rules, to which we add a tenth rule assigning to the "Normal" class if none of the previous rules is satisfied.

\begin{center}
\begin{table}[!htbp]
\caption{\label{rules}Classification rules induced by the Dominance-Based Rough Set Approach}
\resizebox{\columnwidth}{!}{

  \begin{tabular}{|lll|}
 \cellcolor[gray]{0.8} \textbf{Rule}& \cellcolor[gray]{0.8} \textbf{Logical expression} & \cellcolor[gray]{0.8} \textbf{Support}\\
  \hline
Rule 1.& If (Sitting=0.06) \& (Season=-0.33) Then (Output=O)&   9.09\%\\
\hline
Rule 2.& If (Sitting=0.25) \& (Age $\leq$ 0.69) \& (Surgery=1) \& (Disease=1) Then (Output=O)&   9.09\%\\
\hline
Rule 3.& If (Sitting=0.31) \& (Surgery=1) \& (Trauma=0) Then (Output=O)&   9.09\%\\
\hline
Rule 4.& If (Sitting=0.38) \& (Season=1) \& (Alcohol$\leq$0.80) Then (Output=O)&   27.27\%\\
\hline
Rule 5.& If (Sitting=0.44) \& (Season=0.33) Then (Output=O)&   9.09\%\\
\hline
Rule 6.& If (Sitting=0.44) \& (Season=1) \& (Fever=-1) Then (Output=O)&   9.09\%\\
\hline
Rule 7.& If (Sitting=0.50) \& (Disease=0) Then (Output=O)&   18.18\%\\
\hline
Rule 8.& If (Sitting=0.50) \& (Season=0.33) \& (Smoking=-1) \& (Surgery=0) Then (Output=O)&   9.09\%\\
\hline
Rule 9.& If (Sitting=0.88) \& (Fever=-1) Then (Output=O or N)&   100.00\%\\
\hline
Rule 10.& Else (Output=N)&   100.00\%\\
\hline
\end{tabular}

}

\end{table}
\end{center}

\begin{center}
\begin{table}[!htbp]

\caption{\label{results}Confusion Matrix and performance indicators}
\begin{center}

  \begin{tabular}{|l|llll|}

\multicolumn{1}{c|}{}& \cellcolor[gray]{0.8}\textbf{MLP }\cite{gil}& \cellcolor[gray]{0.8}\textbf{SVM} \cite{gil}& \cellcolor[gray]{0.8}\textbf{DT} \cite{gil}& \cellcolor[gray]{0.8}\textbf{DRSA}\\
  \hline
TP&	80&	83&	82&	88\\
  \hline
TN&	6&	3&	2&	12\\
  \hline
FP&	9&	12&	13&	1\\
  \hline
FN&	5&	2&	3&	1\\
  \hline	
    \hline
Accuracy (\%)&	86.00&	86.00&	84.00&	98.03 \\
  \hline
Sensitivity (\%)&	94.11&	97.64&	96.47&	98.87 \\
  \hline
Specificity (\%)&	40.00&	20.00&	13.33&	92.30\\
  \hline
Positive Predictive Value (\%)&	89.88&	87.36&	86.31&	98.87\\
  \hline
Negative Predictive Value&	54.54&	60.00&	40.00&	92.30\\
  \hline

\end{tabular}

\end{center}
\hspace*{-3.5cm}

\end{table}

\end{center}

\begin{center}
\begin{figure}[!htbp]
\begin{subfigure}[b]{0.45\textwidth}    
\caption{\label{accuracy}Accuracy}
\includegraphics[width=\textwidth]{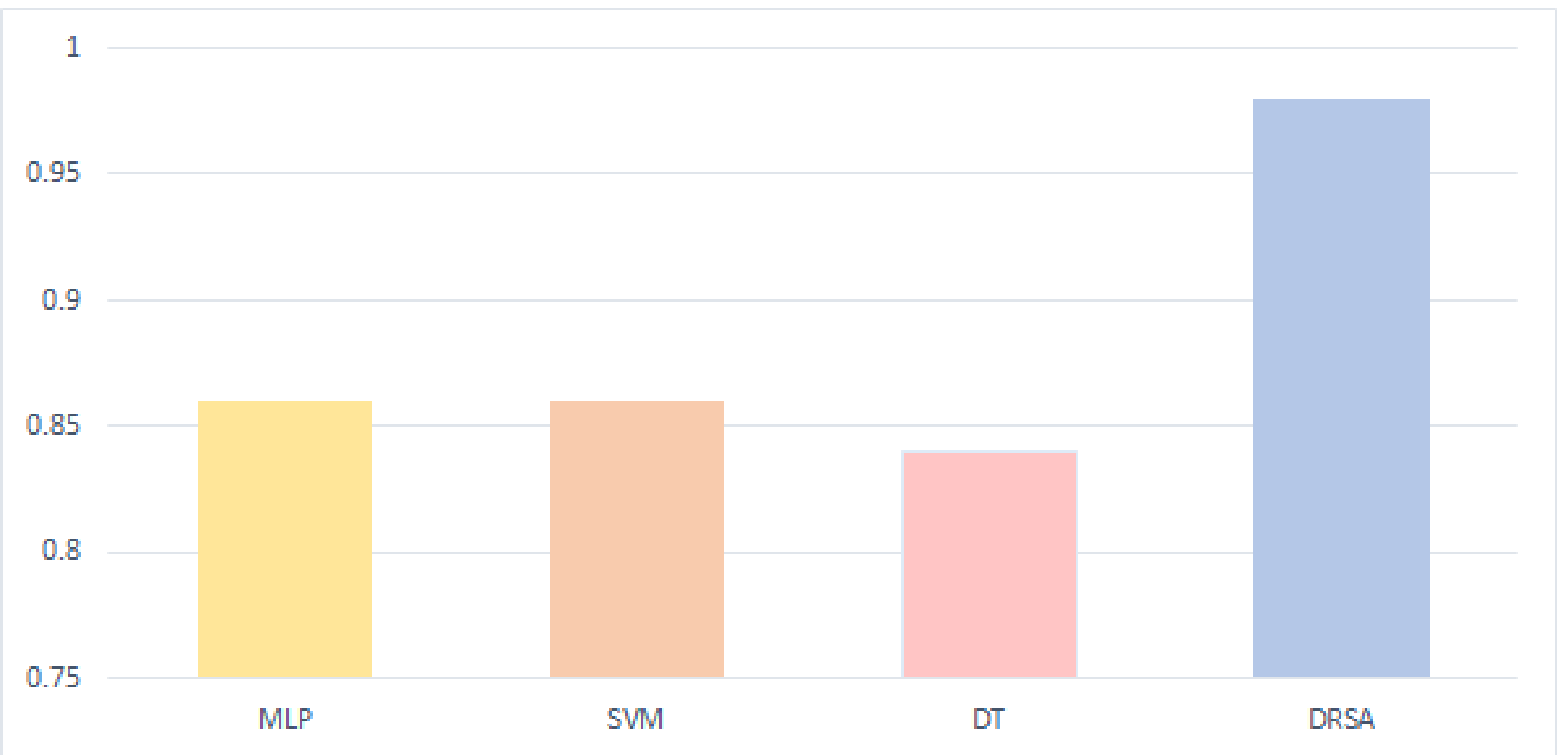}

\end{subfigure}
\quad
\begin{subfigure}[b]{0.45\textwidth}    
\caption{\label{sensitivity}Sensitivity}
\includegraphics[width=\textwidth]{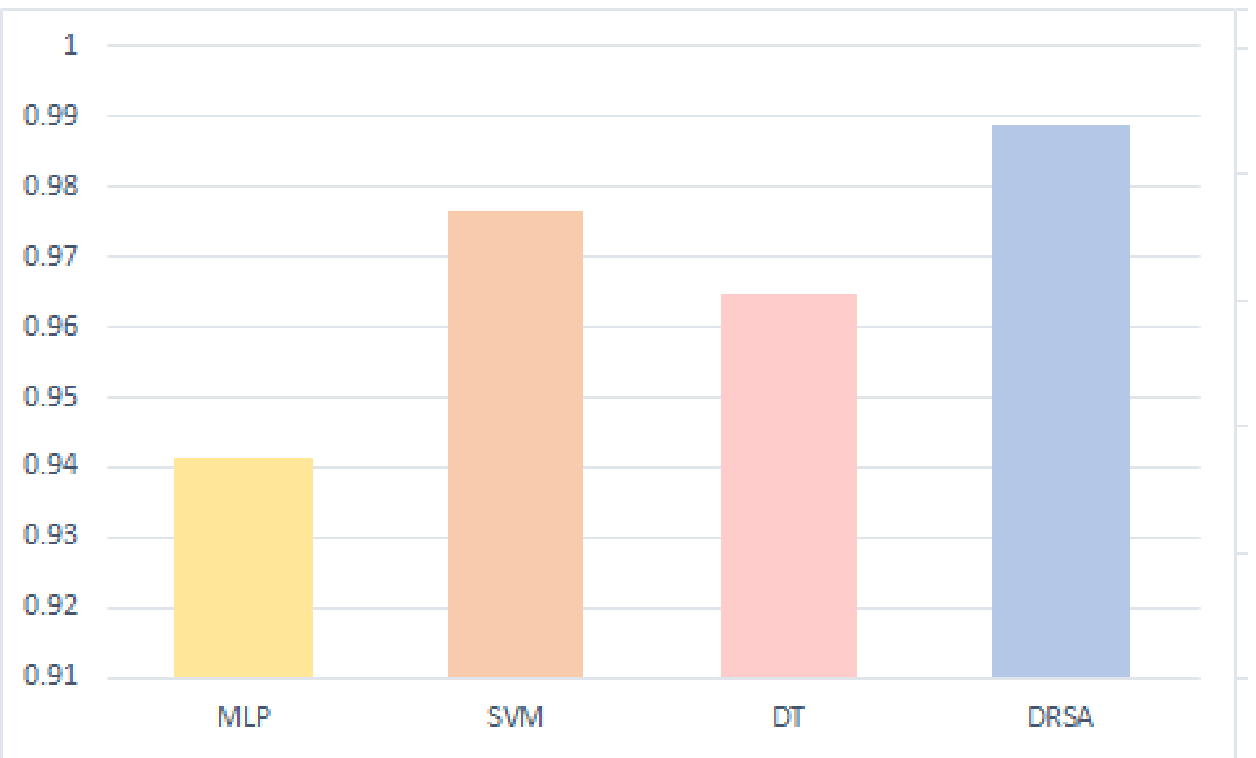}

\end{subfigure}

\begin{subfigure}[b]{0.45\textwidth}    
\caption{\label{specificity}Specificity}
\includegraphics[width=\textwidth]{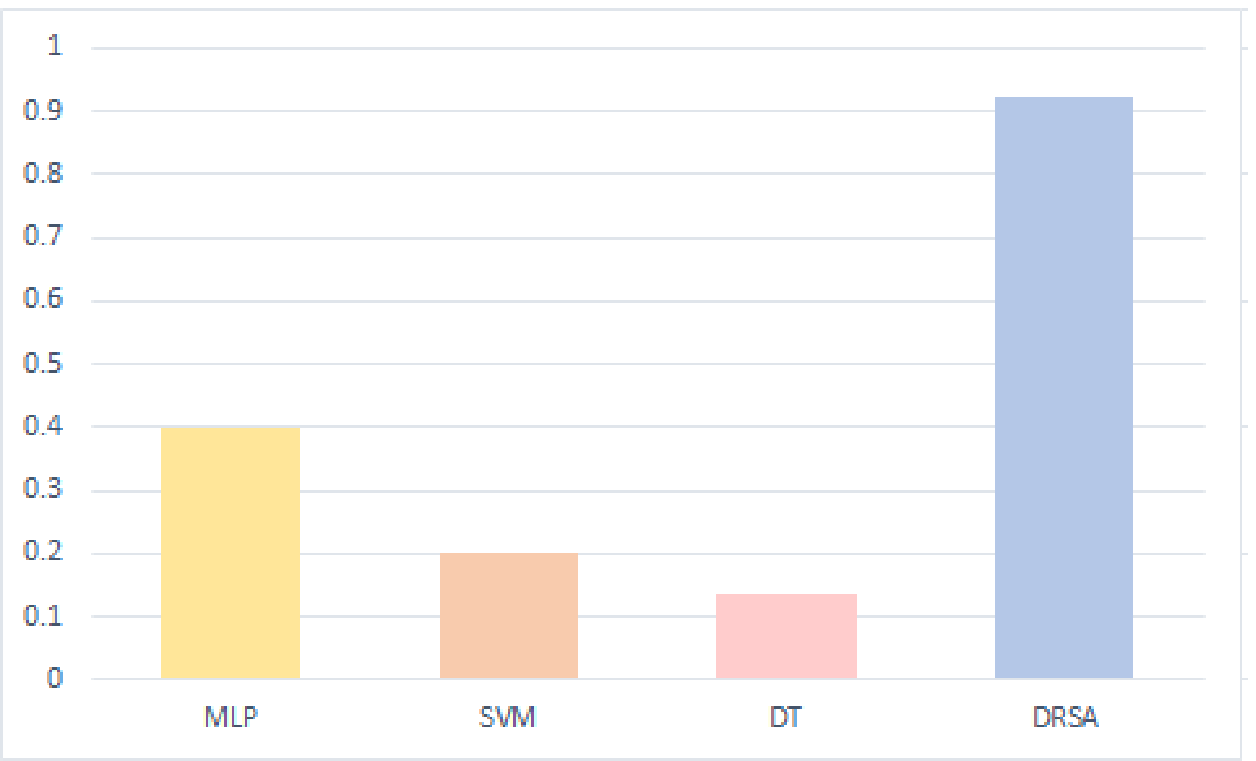}

\end{subfigure}
\quad
\begin{subfigure}[b]{0.45\textwidth}    
\caption{\label{npv}Negative Predictive Value}
\includegraphics[width=\textwidth]{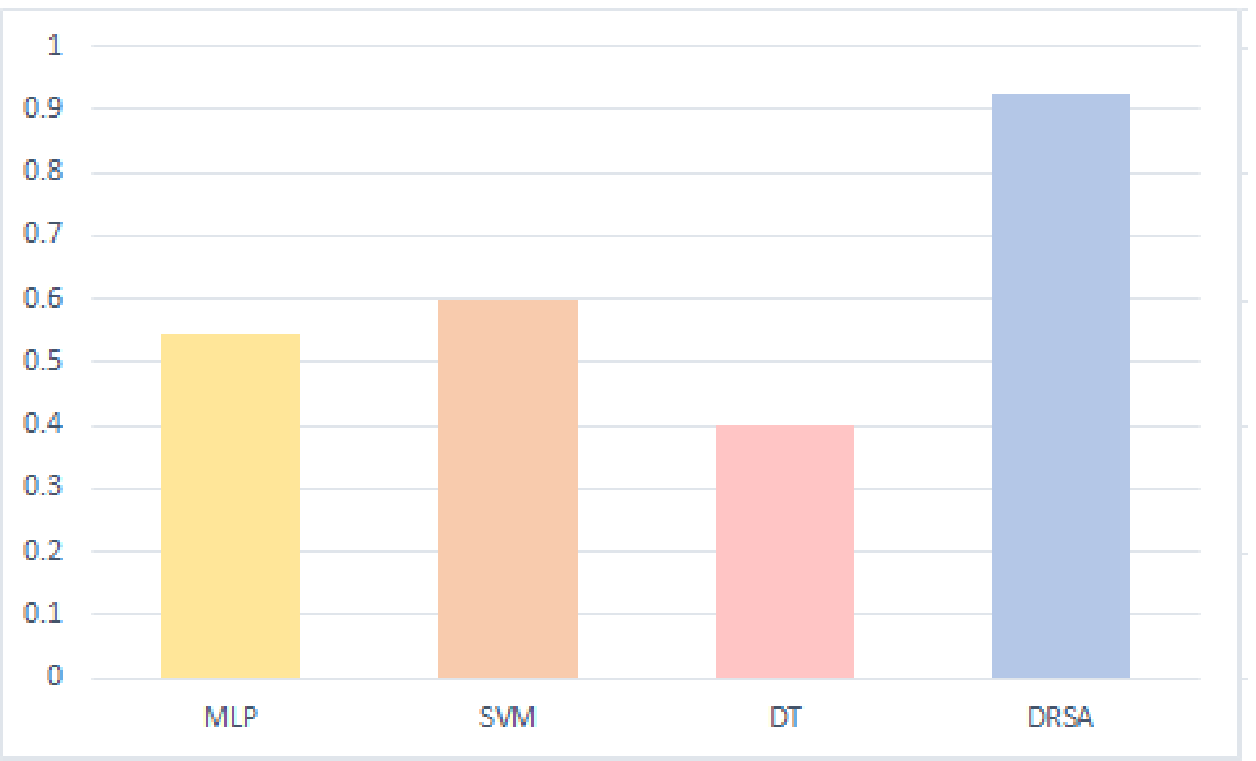}

\end{subfigure}

\begin{subfigure}[b]{0.45\textwidth}    
\caption{\label{ppv}Positive Predictive Value}
\includegraphics[width=\textwidth]{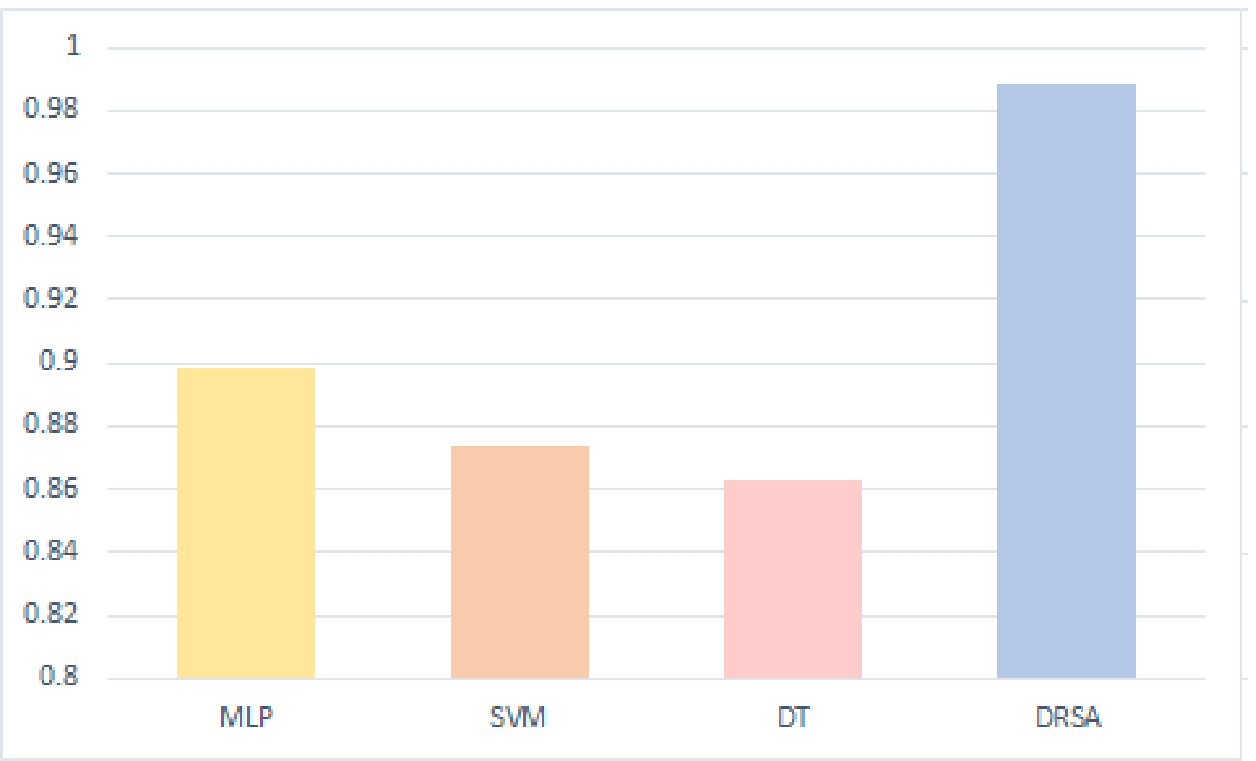}

\end{subfigure}
\caption{\label{figresults}Performance indicators}
\end{figure}

\end{center}

\section{Results and Discussion}
Table \ref{results} and Figure \ref{figresults} compare the performance metrics of this classification system (DRSA) to those obtained in \cite{gil}. As can be observed, the dataset of 100 cases can be described almost exhaustively ($98\%$ accuracy) by the set of ten rules presented in Table \ref{rules}. Further, the $2\%$ inaccurate classification results from an inconsistency in the original dataset, where cases number $67$ and $71$ present the exact same attribute values but are part of two different classes. The metrics achieved by DRSA in Table \ref{results} and Figure \ref{figresults} are thus the highest possible for this dataset and rather than indicating the performance of this approach, they clearly indicate the triviality of  the original clinical dataset of \cite{gil}, which was somehow hidden by the sub-optimal results obtained by methods MLP, SVM, DT in the original publication.\\

Despite the triviality of the reference dataset \cite{UCI} and its small size of 100 instances, conclusions have been drawn by past studies concerning not only the relative technical merits of different machine learning methods \cite{gil}, but also on medical aspects such as the importance of lifestyle factors on male fertility \cite{zhou}. Thus, we insist on the importance of the representativeness of data in any machine learning endeavor and call for the development of objective statistical standards concerning the quality of datasets from which technical and medical conclusions can be drawn. As our results highlight, algorithmic accuracy indicators not only do not reflect the quality of datasets but can more worryingly hide the poor quality of some datasets with high but sub-optimal values. 

\section{Conclusion}
In this research, the dominance-based rough sets approach was utilized on a widely studied reference dataset from the UCI machine learning repository. Due to the monotonic nature of the considered features, the proposed algorithm unsurprisingly outperformed previous machine learning approaches and highlighted serious issues with the quality and representativeness of the reference dataset. Although, there exist elaborate statistical indicators for the performance of machine learning methods on particular datasets, our results suggest a gap in the literature concerning the nonexistence of objective standards for the measurement of data quality ex-ante, that is before a particular machine learning approach is considered. For instance, the explicit definition of conditions to be satisfied by the data (see for instance, condition 1 to 4 of \cite{icic} in the context of natural language processing) seems to be a good practice that would warrant extension to medical diagnosis datasets. 
%% References


\end{document}